\documentclass[pdflatex,sn-mathphys-num]{sn-jnl}

\usepackage{graphicx}%
\usepackage{multirow}%
\usepackage{array} 
\usepackage{booktabs}
\usepackage{amsmath,amssymb,amsfonts}%
\usepackage{amsthm}%
\usepackage{mathrsfs}%
\usepackage[title]{appendix}%
\usepackage{xcolor}%
\usepackage{textcomp}%
\usepackage{manyfoot}%
\usepackage{booktabs}%
\usepackage{listings}%

\usepackage{algorithm}
\usepackage{algpseudocode}

\usepackage{tablefootnote}
\usepackage{hyperref}
\usepackage{subcaption}

\usepackage[algo2e,ruled,vlined]{algorithm2e}

\theoremstyle{thmstyleone}%
\theoremstyle{thmstyletwo}%

\theoremstyle{thmstylethree}%

\begin{document}

\title[Article Title]{DS-SAC: Density Search for Sample Consensus}


\author*{\fnm{Suraj} \sur{Thapa}}\email{sthap18@unh.newhaven.edu}

\author*{\fnm{Muhammad} \sur{Aminul Islam}}\email{MIslam@newhaven.edu}


\affil{\orgdiv{Tagliatela College of Engineering}, \orgname{University of New Haven}, \orgaddress{\street{300 Boston Post Rd}, \city{West Haven}, \postcode{06516}, \state{CT}, \country{USA}}}




\abstract{

Robust geometric model estimation is a fundamental problem in computer vision. RANSAC and its variants remain widely used for this task; however, they rely on stochastic minimal sampling. In this article, we propose \emph{Density Search Sample Consensus} (DS-SAC), a deterministic robust estimation framework, that avoids repeated random sampling by searching dense regions. Starting from an initial model estimated from the available points, the method performs local exploration via forward and backward search. To facilitate global exploration, DS-SAC recursively partitions the point set using signed residuals and searches each valid partition for high-consensus models. We show that DS-SAC has polynomial complexity with respect to the number of points, making it an efficient alternative to stochastic consensus-based methods. Experiments on large-scale real-world datasets for homography, fundamental matrix, and essential matrix estimation show that DS-SAC achieves higher AUC scores, competitive or lower median pose errors, and faster runtime compared with widely used robust estimators, including RANSAC, MAGSAC, LO-RANSAC, and GC-RANSAC.
}
\keywords{RANSAC, Robust Estimation, Optimization}


\maketitle

\section{Introduction}\label{sec1}

Robust model estimation is a fundamental problem in computer vision, where geometric models must be estimated from data containing noise, mismatches, and outliers. This problem appears in many vision tasks, including feature matching, image stitching, relative pose estimation, 3D reconstruction, visual localization, and simultaneous localization and mapping. In these applications, the input points obtained from local feature matching are often contaminated by incorrect matches caused by repetitive structures, illumination changes, occlusions, viewpoint variation, and sensor noise. Therefore, the ability to estimate an accurate model in the presence of a large number of outliers remains a central requirement for reliable geometric vision systems.

A widely used formulation for robust estimation is the maximal consensus problem, where the goal is to find the model that maximizes the number of data points whose residuals lie below a predefined inlier threshold. Random Sample Consensus (RANSAC)~\cite{fischler1981random} is the most influential approach for solving this problem. RANSAC repeatedly samples minimal subsets, estimates candidate models, and selects the model with the largest inlier support. Due to its simplicity and generality, RANSAC has become the standard robust estimator for geometric model fitting. However, its performance depends strongly on the inlier ratio, the minimal sample size, and the inlier threshold. When the outlier ratio is high or the model requires a larger minimal sample, the number of iterations required to obtain a reliable all-inlier sample increases rapidly.

Several RANSAC variants have been proposed to improve sampling efficiency, scoring, and model refinement. MLESAC~\cite{torr2000mlesac} replaces simple inlier counting with a likelihood-based objective, while MSAC~\cite{torr2000mlesac} uses a truncated residual score to distinguish models with the same consensus size. PROSAC~\cite{Chum2005MatchingWP} improves sampling by using prior match quality, and LO-RANSAC~\cite{loransac} refines promising hypotheses using local optimization on the inlier set. GC-RANSAC~\cite{barath2017graphcutransac} further improves local optimization by incorporating spatial coherence through graph-cut based refinement. MAGSAC and MAGSAC++~\cite{barath2019magsacmarginalizingsampleconsensus,barath2019magsacfastreliableaccurate} reduce sensitivity to the manually selected noise scale by marginalizing over a range of thresholds. Although these methods significantly improve the original RANSAC framework, they still rely on random sampling and may require many iterations under high outlier ratios.

Another line of work focuses on deterministic consensus maximization. These methods aim to reduce or remove the stochastic nature of RANSAC by directly searching the model space for high-consensus solutions. Examples include branch-and-bound methods, tree-search strategies, convex or biconvex reformulations, and other deterministic optimization frameworks~\cite{chin2015efficient,cai2018deterministic,le2017exact,7298855,li2009consensus,unitnorm}. Such approaches provide a more systematic search of the solution space and can offer stronger guarantees than random sampling. However, they often suffer from high computational complexity, problem-specific relaxations, heuristic pruning, or approximations that limit their practicality for large-scale image matching.



In this article, we propose \emph{Density Search Sample Consensus} (DS-SAC), a deterministic robust estimation method that searches for dense regions in the residual space instead of relying on random minimal sampling. Rather than repeatedly sampling small subsets, DS-SAC starts by fitting a model using the available points and progressively refines it through forward and backward search. In the forward search, the method gradually reduces the percentile of selected points, allowing the model to move toward a denser and more reliable region of the residual space. In the backward search, the selected region is expanded from the best location found in the forward search. 
To facilitate global exploration, DS-SAC recursively partitions the point set using signed residuals and searches each valid partition for high-consensus models. The best model is selected using the inlier count, with the MSAC score used as a tie-breaking criterion. Compared with RANSAC-based methods, DS-SAC avoids repeated random sampling, explores dense residual regions deterministically, and achieves improved accuracy and runtime across homography, fundamental matrix, and essential matrix estimation.

The source code of our method will be made publicly available upon publication to facilitate reproducibility and future research.

\section{Related Work}\label{sec2}

\subsection{Local Optimization in RANSAC}

Local optimization is commonly used to improve promising RANSAC hypotheses. LO-RANSAC~\cite{loransac} applies an additional refinement step to the best model found so far by re-estimating the model from its inlier set. This improves model accuracy and often increases the final consensus size. Later methods extend this idea using improved sampling, repeated least-squares fitting, or spatially coherent refinement. GC-RANSAC~\cite{barath2017graphcutransac}, for example, incorporates graph-cut optimization to encourage spatial consistency in the inlier labeling.

DS-SAC is related to local optimization because it repeatedly re-estimates models from residual-selected subsets. However, DS-SAC differs from LO-RANSAC and GC-RANSAC in two important ways. First, it does not wait for a random hypothesis to trigger local optimization; instead, the entire search process is built around deterministic residual-density refinement. Second, it uses a forward search and backward search strategy to explore different support sizes, rather than refining only the current consensus set.

\subsection{Deterministic Consensus Maximization}

Deterministic robust estimation methods attempt to solve the consensus maximization problem without relying on random sampling. Branch-and-bound, tree-search, and global optimization methods have been proposed for this purpose~\cite{chin2015efficient,li2009consensus,7298855}. Other works use convex or biconvex relaxations to make the consensus objective more tractable~\cite{cai2018deterministic,le2017exact}. These approaches reduce the stochastic behavior of RANSAC and can provide more systematic exploration of the model space.

Despite their advantages, deterministic methods often require expensive search procedures, problem-specific relaxations, or approximations of the objective and constraints. As a result, they may be difficult to apply efficiently to large point sets. DS-SAC follows the deterministic motivation of these methods but avoids explicit exhaustive search over the full model space. Instead, it searches dense regions in the residual space using percentile-based refinement and recursive signed-residual partitioning. This design provides a practical balance between deterministic exploration, computational efficiency, and robust model accuracy.


The proposed DS-SAC method is positioned between stochastic RANSAC variants and expensive deterministic consensus maximization methods. 
Unlike RANSAC, it does not rely on random minimal sampling. Like deterministic methods, it explores the solution space systematically. However, instead of using branch-and-bound or convex relaxations, DS-SAC uses residual search and recursive partitioning to identify high-consensus regions efficiently. This makes DS-SAC especially efficient for geometric estimation problems such as homography, fundamental matrix, and essential matrix estimation, where point sets may contain many outliers but also exhibit dense inlier structures. By combining deterministic search, local residual-based refinement, MSAC-based tie-breaking, and model-specific solvers, DS-SAC achieves improved robustness and efficiency compared with commonly used RANSAC-based estimators.

\section{Method}\label{sec3}

In this section, we provide an algorithm for our method. Let
\[
\mathcal{X}=\{x_i\}_{i=1}^{N}, \qquad x_i\in\mathbb{R}^{M},
\]
be the full set of input points, where $N=|\mathcal{X}|$. 
Let $\theta\in\mathbb{R}^{Q}$ denote the parameters of a model, and let
\[
d:\mathcal{X}\times\mathbb{R}^{Q}\rightarrow\mathbb{R}_{\geq 0}
\]
be a distance function that measures the fitting error of a point with respect to the model. 
Given an inlier-outlier threshold $T$, the binary inlier function is defined as
\[
\ell_i(\theta)=
\begin{cases}
1, & \text{if } d(x_i,\theta)\le T,\\
0, & \text{otherwise}.
\end{cases}
\]
Thus, a point is labeled as an inlier when its distance from the model is not greater than the threshold, and as an outlier otherwise. The inlier count of a model $\theta$ is then given by
\[
I(\theta)=\sum_{i=1}^{N}\ell_i(\theta).
\]
The maximal consensus problem is to find the model parameters in which maximum number of points are consistent with the model. In other words, the objective is to find the maximum number of inliers.
\[
\theta^*
=
\arg\max_{\theta} I(\theta)
=
\arg\max_{\theta}
\sum_{i=1}^{N}
\ell_i(\theta).
\]

Solving this objective exactly requires an exhaustive search over possible inlier subsets or model hypotheses, which becomes computationally intractable for large $N$. RANSAC and its variants provide approximate solutions by randomly sampling minimal subsets and selecting the model with high consensus. However, their computational cost increases rapidly as the outlier ratio grows, since the probability of sampling an outlier-free subset decreases significantly. In contrast, our method avoids repeated random sampling and searches the residual space deterministically. Its complexity is polynomial with respect to the number of points. 




\begin{algorithm}[t]
\caption{High-Level Overview of DS-SAC}
\label{alg:ds_sac_overview}

\KwIn{A set of tentative points}
\KwOut{Best geometric model and its inlier set}

\BlankLine


\BlankLine

\While{minimum partition size is reached}{

    - {Initialize} a model with all points in the current search space

    - Perform \textbf{forward search} by gradually reducing the selected percentage of points and re-estimating the model\;
    
    
    
    - Store the best model found during the forward search\;
    
    - Perform \textbf{backward search} from the local best model from forward search by gradually increasing the selected percentage of points and re-estimating the model\;
    
    
    - Update the best model if a better consensus is found\;
    
    - \textbf{Partition} the current points into two regions bases on the sign of the residuals\;
    
    - Recursively repeat the same search process on each valid partition\;
}

\BlankLine

Apply final \textbf{post-tuning} to the best model found from recursive partitioning and  search\;


\Return{Final model and inlier set}\;

\end{algorithm}

Our algorithm is based on the observation that inliers form a dense region in the residual space of a good model. 
Starting from an initial model estimate, the method gradually reduces the kernel size, i.e., the distance threshold for selecting points for optimization.  
By gradually decreasing the kernel, the algorithm guides the model towards an increasingly dense residual region. A subsequent \emph{expand} operation gradually increases the kernel size again for local exploration. 
To enable global exploration, the search space is partitioned into two regions and then each of these regions is searched in the aforementioned manner. This search and partition process continues until a minimum partition size is reached. 

Algorithm~\ref{alg:ds_sac_overview} provides a high-level overview of the proposed method.  The algorithm consists of four main components: (i) forward search, (ii) backward search, (iii) recursive partitioning, and (iv) post-tuning. Each component is described in detail in the following subsections. 

\begin{table}[t]
\centering
\caption{Notations.}
\label{tab:notation}
\renewcommand{\arraystretch}{1.15}
\begin{tabular}{p{0.22\linewidth}p{0.68\linewidth}}
\toprule
\textbf{Symbol} & \textbf{Description} \\
\midrule
$\mathcal{X}$ & Full set of input points or correspondences. \\
$N$ & Number of points in $\mathcal{X}$. \\
$\theta$ & Model parameter. \\
$x_i$ & The $i$-th point or correspondence. \\
$\mathcal{S}$ & Set of points in current partition. \\
$p_{\mathrm{partition}}$ & Relative size of the current partition in percentiles, defined as $|\mathcal{S}|/N$. \\
$p_{\min}$ & Minimum allowed partition size in percentiles. \\
$\Delta p$ & Percentile step size used in forward and backward search. \\
$p_{\mathrm{bwd}}$ & Maximum percentile used in the backward search, relative to the current partition size, e.g., $0.5p_{\mathrm{partition}}$. \\

$T$ & Inlier threshold used to classify points as inliers or outliers. \\
$M_{\min}$ & Minimum number of points required for model estimation. \\
$d(x_i,\theta)$ & Distance function used to compute distance of a point $x_i$ with respect to model $\theta$, e. g., Sampson distance for fundamental matrix. \\
$r(x_i, \theta)$ & Signed residual function used for splitting the point set during recursive partitioning. \\
$s(\theta)$ & Model scoring function for evaluation and selection of a model, e. g., inlier count and MSAC score \cite{TORR2000138}. \\

$\theta^*$ & Global best model. \\
$score^*$ & Global best score. \\
\bottomrule
\end{tabular}
\end{table}

\subsection{Forward Search}
Given a partition space $\mathcal{S}\subseteq\mathcal{P}$, an algorithm uses all points in $\mathcal{S}$ to estimate an initial model. This is done by minimizing the sum of the squared distances:
\[
\theta_{\mathrm{init}}
=
\arg\min_{\theta}
\sum_{x_i\in\mathcal{S}} d^2(x_i,\theta).
\]
This initial estimate provides the starting point for the subsequent residual-based forward and backward search. 



%

Let the kernel size for $S$, i.e., the minimum distance threshold within which all points in $S$ lie, be $th$ and the partition size in terms of percentiles of points be $p_0=\frac{|\mathcal{S}|}{N}$.
Starting from this value, the forward search gradually shrinks the kernel size via reducing the percentiles of points by $\Delta p$ until it reaches the minimum partition size, $p_{\min}$.
\[
p_0, p_1=p_0 - \Delta p,  \cdots,  p_K = p_0 - k\Delta p ,\cdots,  p_{\min}.
\]
At each $p_i$, two optimizations are performed. 


\paragraph{Percentile points optimization:}
Suppose the current model is $\theta_c$. We first compute the distance of every point to the model, $d(x_i,\theta_c) \forall x_i\in\mathcal{X}$.
The distances are then sorted in ascending order. Let $th$ be the $\lceil p_iN\rceil$-th smallest distance value. The support set below this threshold is selected as
\[
B=
\left\{
x_i \in \mathcal{X}
\mid
d(x_i,\theta_c)\le th
\right\}.
\]
Thus, $B$ contains the points that are closest to the current model according to the percentile of points, $p_i$. A new candidate model is then estimated from this selected support set.

\paragraph{Inlier optimization:} After each percentile-based model update, we apply an inlier optimization step. For the current model $\theta_c$, the algorithm forms a local support set using the inlier threshold $T$.
If the number of selected inliers is smaller than the minimum number of points required by the solver, the support set is enlarged by including the closest minimum number of points according to their distances. 
The model is then re-estimated from the local support set,
and the locally optimized model is kept only if it improves the current model.

This procedure makes the forward search both selective and adaptive. As the percentile decreases, the model moves toward denser residual regions. However, because residuals are recomputed after every model update, points are not permanently pruned from the full search process; a point may be selected again if it becomes consistent with an updated model.

\begin{algorithm}
\caption{Forward Search}
\label{alg:ds1}
\SetAlgoLined
\KwIn{
Set of points in the current partition, $\mathcal{S}$ 
}
Partition size, $p_{partition} \leftarrow |\mathcal{S}|/N$

\tcc{Initialization}
Initial parameters, $\theta_{c} \leftarrow \arg\min_\theta \sum_{x_i \in \mathcal{S}}d(x_i,\theta)^2$

$\theta_{best,local} \leftarrow \theta_{c}$

$score_{best, local} \leftarrow s(\theta_{c})$

$p_{best} \leftarrow p_{partition}$

\tcc{Forward Search}
\For{$p$ in $p_{partition}$:$-\Delta p$:$p_{min}$}{
\tcc{Percentile points optimization}
$d_i \leftarrow d(x_i, \theta_c), \forall x_i \in P$

$th \leftarrow \lceil pN\rceil$-th smallest $d_i$

$B \leftarrow \{x_i | d(x_i, \theta) \leq th\}$

\refstepcounter{equation}\label{eq:theta-local-update}%
$\displaystyle \theta_c \leftarrow \arg\min_{\theta} \sum_{x_i \in B} d(x_i,\theta)^2 $ \hfill(\theequation)\;


$score_c \leftarrow s(\theta_c)$

Update $score_{best, local}$, $\theta_{best,local}$ and $p_{best}$ if $score_{best, local} > score_c$.

\tcc{Inlier optimization}
$d_i \leftarrow d(x_i, \theta_c), \forall x_i \in P$

Inliers, $B \leftarrow \{x_i | d(x_i, \theta) \le T\}$

\If{$ |B| < M_{min}$}{

$B \leftarrow$ top $M_{min}$ points with ascending $d_i$ 
}
\refstepcounter{equation}\label{eq:inlier_opt}%
$ \displaystyle \theta_{I} \leftarrow \arg\min_{\theta} \sum_{x_i \in B} d(x_i,\theta)^2$ \hfill(\theequation)\;

$score_I \leftarrow s(\theta_I)$

\If{$ score_I > score_c$}{
$\theta_c \leftarrow \theta_I$

}
Update $score_{best, local}$, $\theta_{best,local}$ and $p_{best}$ if $score_{best, local} > score_I$.

}

\KwResult{$score_{best, local}$, $\theta_{best,local}$, $p_{best}$}
\end{algorithm}

\subsection{Backward Search}

\begin{algorithm}
\caption{Backward Search}
\label{alg:bwd-search}
\SetAlgoLined
\KwIn{
Set of points $\mathcal{S}$, best forward score $score_{best, local}$, best forward parameter $\theta_{best,local}$, best percentile $p_{best}$
}

$\theta_{c} \leftarrow \theta_{best,local}$


\tcc{Backward Search}
\For{$p$ in $p_{best}$:$\Delta p$:$p_{bwd}$}{
\tcc{Percentile points optimization}
$d_i \leftarrow d(x_i, \theta_c), \forall x_i \in P$

$th \leftarrow \lceil pN\rceil$-th smallest $d_i$

$B \leftarrow \{x_i | d(x_i, \theta) \leq th\}$

$\theta_c \leftarrow \arg\min_{\theta} \sum_{x_i \in B} d(x_i,\theta)^2$

$score_c \leftarrow s(\theta_c)$

Update $score_{best, local}$ and $\theta_{best,local}$ if $score_{best, local} > score_c$.

\tcc{Inlier optimization}
$d_i \leftarrow d(x_i, \theta_c), \forall x_i \in P$

Inliers, $B \leftarrow \{x_i | d(x_i, \theta) \le T\}$

\If{$ |B| < M_{min}$}{

$B \leftarrow$ top $M_{min}$ points with ascending $d_i$ 
}

$\theta_{I} \leftarrow \arg\min_{\theta} \sum_{x_i \in B} d(x_i,\theta)^2$

$score_I \leftarrow s(\theta_I)$

\If{$ score_I > score_c$}{
$\theta_c \leftarrow \theta_I$

}
Update $score_{best, local}$ and $\theta_{best,local}$ if $score_{best, local} > score_I$.

}

\KwResult{$score_{best, local}$, $\theta_{best,local}$}
\end{algorithm}



Following the forward search, backward search is performed in the neighborhood regions of the best parameters found by gradually increasing the kernel size.  
The algorithm is similar to forward search, but in reverse direction with the initial parameter as the local best parameter from the forward search and the start percentile is the best percentile from the forward search. However, the maximum percentile used in this stage can vary depending on the complexity and type of the problem. This step allows for outward exploration from the dense region found in forward search.

\subsection{Recursive Partitioning}

After the forward and backward search are completed for a partition space, we use the best model obtained from the forward search as the basis for splitting the current space into two regions. The points in the current partition are divided according to the sign of the signed residual:
\[
\mathcal{S}^{+}= \{ x_i | x_i \in S, r(x_i, \theta_{split})>0\}
\]
\[
\text{and }\mathcal{S}^{-}= \{ x_i | x_i \in S, r(x_i, \theta_{split})\leq 0\},
\]
where $r_i(\theta)$ is a signed residual function. 


The partition defined by the parameter, $\theta_{split}$, must lie within the current search space to ensure a signed-based split always produces two regions. 
In the case when the partition boundary is outside the current search space--because of the denser region being outside--we re-scan the space in the forward direction without the inlier optimization step and use the parameters with the smallest kernel size for which the boundary lies within. In the worst case, $\theta_{init}$ will be the partition parameter. This exception handling case is not shown in the presented Algorithm \ref{alg:recursive} in order to keep the core part of the algorithm uncluttered for clarity.


\begin{algorithm}
\caption{Search and Partition}
\label{alg:recursive}
\SetAlgoLined
\KwIn{
Set of points $\mathcal{S}$

}
Partition size, $p_{partition} \leftarrow |\mathcal{S}|/N$

\If{$p_{partition} < p_{\min}$}{
return
}
\tcc{Search}
($score_{best, local}$, $\theta_{best,local}$, $p_{best}$) $\leftarrow$ \textsc{Forward\_Search}$(\mathcal{S})$  

$\theta_{split} \leftarrow \theta_{best,local}$ 

($score_{best, local}$, $\theta_{best,local}$) $\leftarrow$ \textsc{Backward\_Search}$(\mathcal{S}$,$score_{best, local}$, $\theta_{best,local}$, $p_{best}$)

Update $\theta^*$ and $score*$ if $score_{best, local} > score*$

\tcc{Partitioning}
$\mathcal{S}_1 \leftarrow \{x_i | r_i(\theta_{split}) \geq 0\}$ \tcp*{$r$ is the signed residual function} 

$\mathcal{S}_2 \leftarrow \{x_i | r_i(\theta_{split}) < 0\}$\\

\textsc{Search\_and\_Partition}$(\mathcal{S}_1)$ \tcp*{Recursion, calls the current function} 

\textsc{Search\_and\_Partition}$(\mathcal{S}_2)$ \tcp*{Recursion} 

\BlankLine
\textbf{Results:}
Best model $\theta^*$ and score $score*$.
\end{algorithm}

\subsection{Post-Tuning}
After the recursive search returns the best model, we apply a final post-tuning stage. Starting from the best model $\theta^*$, the method constructs point sets using a sequence of relaxed-to-strict thresholds:
\[
T={k_1T,k_2T,\dots,k_LT},
\]
where $k_i > k_{i+1}$. For each threshold factor $k_i$, the algorithm selects the corresponding support set, re-estimates the model, and evaluates it. A post-tuned model is accepted only if it improves the current best model. 
This final stage improves the geometric accuracy of the model while preserving the robust consensus objective.

\section{Computational Complexity}


The complexity analysis is carried out with a binary tree representation of the partitions. We first perform the worst-case analysis, which occurs when the tree is extremely imbalanced. At each split, one descendant node contains a single point while the remaining points are in other nodes.

The size of the root node is $1$ (as a fraction of the total points).  After partitioning, the size of the minor node is $\frac{1}{N}$ and the major node is   $(1 - \frac{1}{N})$. At depth $k$, the size of the major node is $(1- k\frac{1}{N})$. The partitioning process continues until the major node has exactly one point with a size of $\frac{1}{N}$. That is,
\[1 - k\frac{1}{N} = \frac{1}{N},\]
\[\text{or, } k = N-1.\]
Consequently, the height of the tree is $N-1$. Each step involves $4$ iterations (2 each during forward and backward scans). The total number of iterations expressed as a sum of series over all nodes is
\[c_{total} = \frac{4}{\Delta p} \left(\sum_{k=0}^{N-1} (1 - k\frac{1}{N}) + \sum_{k=1}^{N-2} 1/N \right)\]
where the first and second terms correspond to the major and minor nodes, respectively. Using the smallest possible value, $\frac{1}{N}$, for $\Delta p$ yields, $c_{total} =  2(N^2 + 3N -4)$. As post-tuning adds a constant number of iterations, the algorithm has a complexity of $\mathcal{O}(N^2)$. 

We also analyze the case when the tree is fully balanced with points equally distributed among nodes at each level of depth.
The height of the tree is determined such that each leaf node has just one point and the partition size $1/N$.
\[2^{-k} = 1/N\]
\[\text{or, } k = \log_2(N)\]
As each level requires $\frac{4}{\Delta p}$ iterations, the total number of iterations is 
\[ \frac{4}{\Delta p} \sum_{k=0}^{\log_2N} (1) = \frac{4\log_2 N+4}{\Delta p} \le 4N (\log_2 N + 1).\]
As such, the complexity is $\mathcal{O}(N\log_2N)$.

\section{Experiments and Results}\label{sec4}
In this section, we present a comprehensive evaluation of our method on multiple geometric vision tasks, including fundamental matrix, essential matrix, and homography estimation. The objective is to evaluate our method in terms of pose accuracy, inlier selection quality, and computational efficiency against widely used robust estimators.

\subsection{Datasets}
We evaluate on a diverse collection of indoor and outdoor datasets with different scene structures and ground-truth acquisition procedures. The evaluation includes ScanNet1500~\cite{dai2017scannetrichlyannotated3dreconstructions,sarlin2020supergluelearningfeaturematching}, PhotoTourism~\cite{PhotoTourism,Jin_2020}, LaMAR~\cite{sarlin2022lamarbenchmarkinglocalizationmapping}, 7Scenes~\cite{6671777}, ETH3D~\cite{8954208}, and KITTI~\cite{kitti}. These datasets cover challenging conditions such as wide-baseline outdoor views, indoor RGB-D scenes, driving sequences, and augmented-reality localization scenarios.
The image pairs for evaluation were selected following the dataset setup and evaluation protocol of SupeRANSAC~\cite{barath2025superansac}. For PhotoTourism, we use the 9900 validation pairs from the CVPR Image Matching Challenge 2020. For ScanNet, we use the standard test set consisting of 1500 image pairs. For 7Scenes, image pairs are selected from the test sequences by sampling every 10th image $i$ and pairing it with image $(i+50)$, resulting in 1600 pairs. For ETH3D, we sample pairs from the 13 training scenes that share at least 500 ground-truth keypoints, resulting in 1969 pairs. For KITTI, we form 23,190 image pairs from consecutive frames in the 11 training sequences. For LaMAR, we use 1423 consecutive image pairs from the HoloLens validation split of the CAB scene. In total, the evaluation covers 39,592 image pairs across indoor, outdoor, driving, and augmented-reality scenarios.
The final reported scores are averaged across datasets so that each dataset contributes equally to the overall performance, regardless of the number of image pairs. 

We follow the evaluation protocol of the Image Matching Challenge (IMC) benchmark~\cite{Jin_2020} for epipolar geometry estimation. The main evaluation metric is the relative pose error, defined as the maximum of the rotation error and translation error between the estimated and ground-truth camera poses. Rotation error is computed as the angular difference between the estimated and ground-truth rotation matrices, while translation error is measured as the angle between the estimated and ground-truth translation directions. To summarize performance across different error thresholds, we report the Area Under the recall Curve (AUC)~\cite{AUC} at $5^\circ$, $10^\circ$, and $20^\circ$. Higher AUC values indicate that a method recovers accurate camera poses for a larger fraction of image pairs.

\subsection{Experimental Setup} 
Data normalization is applied before model estimation to improve numerical stability. For homography and fundamental matrix estimation, where the input correspondences are given in image coordinates, we normalize the points by translating their centroid to the origin and scaling them so that the average distance from the origin is $\sqrt{2}$, following Hartley~\cite{601246}. For essential matrix estimation, where camera intrinsics are available, the image points are first transformed into normalized camera coordinates using the corresponding focal lengths and principal points. 

For all experiments, sparse feature points are obtained using SuperPoint~\cite{detone2018superpointselfsupervisedpointdetection} features matched with LightGlue~\cite{lindenberger2023lightgluelocalfeaturematching}. DS-SAC is compared against several representative robust estimation baselines, including OpenCV RANSAC, MAGSAC~\cite{barath2019magsacmarginalizingsampleconsensus}, LO-RANSAC~\cite{loransac}, and GC-RANSAC~\cite{barath2017graphcutransac}. All methods are evaluated using a fixed budget of 1000 iterations to ensure a fair comparison. 

For model scoring, DS-SAC primarily follows the maximal consensus criterion by selecting the model with the highest inlier count. When two candidate models produce the same number of inliers, their performance is considered equal under the consensus objective. In this case, we use the MSAC score as a secondary criterion to distinguish between them. The MSAC score is defined as \[ C_{\mathrm{MSAC}}(\theta)= \sum_{i=1}^{N} \max\left(1-\frac{d^2(x_i,\theta)}{T^2},0\right). \] This score assigns larger values to correspondences with smaller residuals and assigns zero contribution to points whose residuals exceed the inlier threshold. Therefore, the final model is selected first by maximizing the inlier count and, in the case of equal inlier counts, by choosing the model with the higher MSAC score. We used a truncated loss with threshold $1.5T$ as used in GC-RANSAC implementation \footnote{https://github.com/danini/graph-cut-ransac}. Our score is a tuple consisting of inlier count and MSAC score. 

The main DS-SAC parameters are kept fixed across the experiments. The robust percentile step size is set to $\Delta=0.03$, and the minimum partition percentage factor is set to $p_{\min}=0.2$. The inlier threshold $T$ was set to $T = 3.84\sigma^2$ for epipolar geometry and $T = 5.99\sigma^2$ for homography estimation. In both cases, the expected noise standard deviation was set to $\sigma = 0.3$. The inlier thresholds are derived from the corresponding Chi-square distribution according to the degrees of freedom of each geometric model. 

All methods are evaluated using the same input points for a fair comparison. For DS-SAC, the solver and residual function are selected according to the geometric model being estimated. To reduce computation time, DS-SAC precomputes reusable linear coefficients for the geometric solvers. The tests were conducted on a workstation equipped with a 16-core CPU (3.0GHz) and $1$ TB of RAM.

\subsection{Homography Estimation}


We also evaluate DS-SAC on homography estimation using the same image pairs. Homography estimation is important for planar scenes or camera motions where a dominant plane explains the image structure. We compare all methods using pose-based AUC, median pose error, average number of inliers, and runtime.

For homography estimation, we use the four-point algorithm. A homography relates corresponding points in homogeneous coordinates as
\[
\mathbf{x}_i' \sim \mathbf{H}\mathbf{x}_i,
\qquad 
\mathbf{H}\in\mathbb{R}^{3\times 3},
\quad \det(\mathbf{H})\neq 0.
\]
Let $\mathbf{x}_i=(u_i,v_i,1)^\top$ and $\mathbf{x}_i'=(u_i',v_i',1)^\top$. Expanding the homography constraint gives two equations per correspondence. We use the following equation as the partition boundary. 
\[
u_i h_{11}+v_i h_{12}+h_{13}
-
u_i'(u_i h_{31}+v_i h_{32}+h_{33})
= 0.
\]
In this case, the signed residual function is
\[
r((\mathbf{x}_i, \mathbf{x}_i'), \mathbf{H})
=
u_i h_{11}+v_i h_{12}+h_{13}
-
u_i'(u_i h_{31}+v_i h_{32}+h_{33}).
\]
where the homography matrix is defined as
\[
\mathbf{H}
=
\begin{bmatrix}
h_{11} & h_{12} & h_{13}\\
h_{21} & h_{22} & h_{23}\\
h_{31} & h_{32} & h_{33}
\end{bmatrix}.
\]

We use the reprojection error as the distance and inlier functions. The homography is estimated using the \emph{Direct Linear Transform} (DLT).

\begin{table}[htbp]
  \centering
    \begin{tabular}{l|rrrrrr}
    \toprule
    \multicolumn{1}{c}{Method} & \multicolumn{1}{c}{AUC@$5^\circ$} & \multicolumn{1}{c}{AUC@$10^\circ$} & \multicolumn{1}{c}{AUC@$20^\circ$} & \multicolumn{1}{c}{med. $\epsilon$} & \multicolumn{1}{c}{\# inliers } & \multicolumn{1}{c}{time (s)} \\
    \midrule
    RANSAC [OpenCV] \tablefootnote{OpenCV RANSAC implementation: \url{https://docs.opencv.org/5.0/}} & 16.98 & 29.85 & 46.01 & 5.53  & 48.43 & 0.012 \\
    MAGSAC [Author] \tablefootnote{MAGSAC implementation: \url{https://github.com/danini/magsac}} & 16.72 & 29.39 & 45.33 & 5.5   & 48.49 & 0.013 \\
    LO-RANSAC [PoseLib] \tablefootnote{LO-RANSAC implementation from PoseLib: \url{https://github.com/PoseLib/PoseLib}} &22.31 & 34.89 & 49.96 & 5.03  & \textbf{57.99} & 0.009 \\
    GC-RANSAC [Author] \tablefootnote{GC-RANSAC implementation: \url{https://github.com/danini/graph-cut-ransac}} &20.63 & 34.3  & 50.35 & 4.54  & 55.34 & 0.017 \\
    DS-SAC    & \textbf{26.74} & \textbf{40.9} & \textbf{56.38} & \textbf{3.87} & 55.68 & \textbf{0.007} \\

    \bottomrule
    \end{tabular}%
  \caption{Performance comparison for homography estimation across ScanNet1500, PhotoTourism, LaMAR, 7Scenes, ETH3D, and KITTI. Feature points are obtained using SuperPoint+LightGlue. All methods are evaluated with a fixed budget of 1000 iterations. DS-SAC took 465.21 iterations on average. We report the AUC of relative pose error at $5^\circ$, $10^\circ$, and $20^\circ$, median relative pose error in degrees, the average number of inliers, and the average runtime per image pair.}
  \label{tab:homography}
\end{table}

Table~\ref{tab:homography} presents the homography estimation results. DS-SAC achieves the best performance across all AUC thresholds, with $26.74$ at $5^\circ$, $40.90$ at $10^\circ$, and $56.38$ at $20^\circ$. The improvement over the strongest competing method is substantial. At $10^\circ$, DS-SAC improves the AUC from $34.89$ for LO-RANSAC and $34.30$ for GC-RANSAC to $40.90$. At $20^\circ$, DS-SAC also improves over GC-RANSAC from $50.35$ to $56.38$.

DS-SAC also obtains the lowest median pose error of $3.87^\circ$, compared with $4.54^\circ$ for GC-RANSAC and $5.03^\circ$ for LO-RANSAC. In addition, DS-SAC achieves the best runtime of $0.007$ seconds per image pair. Although LO-RANSAC produces the highest average number of inliers, DS-SAC achieves better pose accuracy with a slightly smaller inlier count. This again suggests that the inliers selected by DS-SAC are more geometrically meaningful for estimating the final model. Overall, the homography results show that DS-SAC provides a strong balance between accuracy, robustness, and computational efficiency.

\subsection{Fundamental Matrix Estimation}


The fundamental matrix $\mathbf{F}\in\mathbb{R}^{3\times 3}$ represents the epipolar geometry between two uncalibrated views. For a pair of corresponding homogeneous image points $\mathbf{x}_i$ and $\mathbf{x}_i'$, the fundamental matrix satisfies the epipolar constraint
\begin{equation}
    \mathbf{x}_i'^{\top}\mathbf{F}\mathbf{x}_i = 0.
    \label{eq:epipolar}
\end{equation}

It maps a point in one image to its corresponding epipolar line in the other image and has rank two.

For fundamental matrix estimation, we use the eight-point solver. 
The distance function is defined using the Sampson distance.
\begin{equation}
    d((\mathbf{x}_i,\mathbf{x}_i'), \mathbf{F})
    =
    \frac{(\mathbf{x}_i'^{\top}\mathbf{F}\mathbf{x}_i)^2}
    {(\mathbf{F}\mathbf{x}_i)_1^2+(\mathbf{F}\mathbf{x}_i)_2^2
    +(\mathbf{F}^{\top}\mathbf{x}_i')_1^2+(\mathbf{F}^{\top}\mathbf{x}_i')_2^2}
    \label{eq:dist_func_FM}
\end{equation}
The residual function is derived from Eq. \ref{eq:epipolar}, the epipolar constraint.
\begin{equation}
    r((\mathbf{x}_i,\mathbf{x}_i'), \mathbf{F}) = \mathbf{x}_i'^{\top}\mathbf{F}\mathbf{x}_i
    \label{eq:residual_dist_func_FM}
\end{equation}
We also experimented with the seven-point algorithm \cite{f114e93b39e045d8ad0339824d1625c5}; however, it did not improve the overall performance, so we report the results using the eight-point algorithm which is comparatively computationally efficient.
Bundle optimization is applied to further refine the estimated fundamental matrix.

\begin{table}[htbp]
  \centering
    \begin{tabular}{l|rrrrrr}
    \toprule
    \multicolumn{1}{c}{Method} & \multicolumn{1}{c}{AUC@$5^\circ$} & \multicolumn{1}{c}{AUC@$10^\circ$} & \multicolumn{1}{c}{AUC@$20^\circ$} & \multicolumn{1}{c}{med. $\epsilon$} & \multicolumn{1}{c}{\# inliers } & \multicolumn{1}{c}{time (s)} \\
    \midrule
    RANSAC [OpenCV] & 30.05 & 44.87 & 59.02 & 3.06  & 255.74 & 0.012 \\
    MAGSAC [Author] & 30.32 & 43.59 & 56.17 & 3.48  & 261.57 & 0.278 \\
    LO-RANSAC [PoseLib] & 35.33 & 50.14 & 63.43 & 2.56  & 268.24 & 0.01 \\
    GC-RANSAC [Author] & 43.12 & 55.42 & 66.38 & 2.29  & \textbf{284.44} & 0.032 \\
    DS-SAC    & \textbf{44.38} & \textbf{57.29} & \textbf{68.68} & \textbf{2.06} & 283.29 & \textbf{0.008} \\
    \bottomrule
    \end{tabular}%
  \caption{Performance comparison for fundamental matrix estimation across ScanNet1500, PhotoTourism, LaMAR, 7Scenes, ETH3D, and KITTI, totaling 39,592 image pairs. Feature points are obtained using SuperPoint+LightGlue. All methods are evaluated with a fixed budget of 1000 iterations. DS-SAC took 490.31 iterations on average. We report the AUC of relative pose error at $5^\circ$, $10^\circ$, and $20^\circ$, median relative pose error in degrees, the average number of inliers, and the average runtime per image pair.}
  \label{tab:fundamental_matrix}
\end{table}

Table~\ref{tab:fundamental_matrix} reports the results for fundamental matrix estimation. DS-SAC achieves the best pose accuracy across all AUC thresholds, obtaining an AUC of $44.38$ at $5^\circ$, $57.29$ at $10^\circ$, and $68.68$ at $20^\circ$. Compared with the strongest baseline, GC-RANSAC, DS-SAC improves the AUC from $55.42$ to $57.29$ at $10^\circ$ and from $66.38$ to $68.68$ at $20^\circ$. DS-SAC also achieves the lowest median pose error of $2.06^\circ$, compared with $2.29^\circ$ for GC-RANSAC and $2.56^\circ$ for LO-RANSAC. This indicates that DS-SAC not only improves recall at larger error thresholds but also produces more accurate poses on typical image pairs.

In terms of inlier selection, GC-RANSAC obtains the highest average number of inliers, while DS-SAC produces a very similar inlier count. However, DS-SAC achieves better pose accuracy despite using slightly fewer inliers, suggesting that its selected inlier set is geometrically more consistent. DS-SAC is also the fastest method in this comparison, with an average runtime of $0.008$ seconds per image pair. This demonstrates that the proposed search strategy improves both accuracy and efficiency for fundamental matrix estimation.


\subsection{Essential Matrix Estimation}


The essential matrix $\mathbf{E}\in\mathbb{R}^{3\times 3}$ represents the epipolar geometry between two calibrated views. For a pair of corresponding normalized image points $\mathbf{x}_i$ and $\mathbf{x}_i'$, it satisfies the epipolar constraint
\[
\mathbf{x}_i'^{\top}\mathbf{E}\mathbf{x}_i = 0.
\]
Unlike the fundamental matrix, the essential matrix assumes known camera intrinsics and relates correspondences in normalized camera coordinates.

The distance function and the signed residual function are the same as those used for fundamental matrix estimation, defined in Eq.~\eqref{eq:dist_func_FM} and Eq.~\eqref{eq:residual_dist_func_FM}, respectively, where $\mathbf{x}$ and $\mathbf{x}'$ are camera normalized co-ordinates, and $\mathbf{F}$ is replaced with $\mathbf{E}$.
For essential matrix estimation, we use the Nist\'er five-point solver~\cite{1288525} for the inlier optimization step and the eight-point algorithm with rank enforcement for percentile point optimization in Algorithms \ref{alg:ds1} and \ref{alg:bwd-search}. The input image points are first transformed into normalized camera coordinates using the intrinsic matrices. The five-point solver operates directly on these camera-normalized correspondences. However, the points are further normalized for the eight-point algorithm using Hartley normalization \cite{601246}.
The inlier function is the Sampson distance. Bundle optimization is also applied to further refine the estimated essential matrix.

\begin{table}[htbp]
  \centering
    \begin{tabular}{l|rrrrrr}
    \toprule
    \multicolumn{1}{c}{Method} & \multicolumn{1}{c}{AUC@$5^\circ$} & \multicolumn{1}{c}{AUC@$10^\circ$} & \multicolumn{1}{c}{AUC@$20^\circ$} & \multicolumn{1}{c}{med. $\epsilon$} & \multicolumn{1}{c}{\# inliers } & \multicolumn{1}{c}{time (s)} \\
    \midrule
    RANSAC [OpenCV] & 42.53 & 58.11 & 70.68 & 1.85  & 257.4 & 0.048 \\
    MAGSAC [Author] & 40.51 & 51.23 & 59.79 & 1.69  & 236.51 & 0.402 \\
    LO-RANSAC [PoseLib] & 41.21 & 54.57 & 66.34 & 2.25  & \textbf{281.8} & 0.03 \\
    GC-RANSAC [Author] & 50.93 & 62.15 & 70.97 & \textbf{1.39} & 271.6 & 0.038 \\
    DS-SAC    & \textbf{51.29} & \textbf{63.87} & \textbf{74.35} & 1.72  & 275.46 & \textbf{0.015} \\
    \bottomrule
    \end{tabular}%
  \caption{Performance comparison for essential matrix estimation across the same six datasets used for fundamental matrix estimation. Feature points are obtained using SuperPoint+LightGlue. All methods are evaluated with a fixed budget of 1000 iterations. DS-SAC took 486 iterations on average. We report the AUC of relative pose error at $5^\circ$, $10^\circ$, and $20^\circ$, median relative pose error in degrees, the average number of inliers, and the average runtime per image pair.}
  \label{tab:essential_matrix}
\end{table}

The results are shown in Table~\ref{tab:essential_matrix}. DS-SAC achieves the highest AUC across all pose error thresholds, with $51.29$ at $5^\circ$, $63.87$ at $10^\circ$, and $74.35$ at $20^\circ$. Compared with GC-RANSAC, the strongest baseline in terms of AUC, DS-SAC improves the AUC from $62.15$ to $63.87$ at $10^\circ$ and from $70.97$ to $74.35$ at $20^\circ$. These improvements show that DS-SAC provides more reliable relative pose estimation, especially at moderate and larger error thresholds.

GC-RANSAC obtains the lowest median pose error, and LO-RANSAC obtains the highest average number of inliers in this experiment. However, DS-SAC achieves the best AUC scores while being substantially faster. DS-SAC requires only $0.015$ seconds per image pair, compared with $0.038$ seconds for GC-RANSAC and $0.402$ seconds for MAGSAC. This result highlights an important advantage of DS-SAC: it provides the best overall recall performance while maintaining a very low computational cost.


\begin{figure}[htbp]
    \centering

    \begin{subfigure}{0.48\linewidth}
        \centering
        \includegraphics[width=\linewidth]{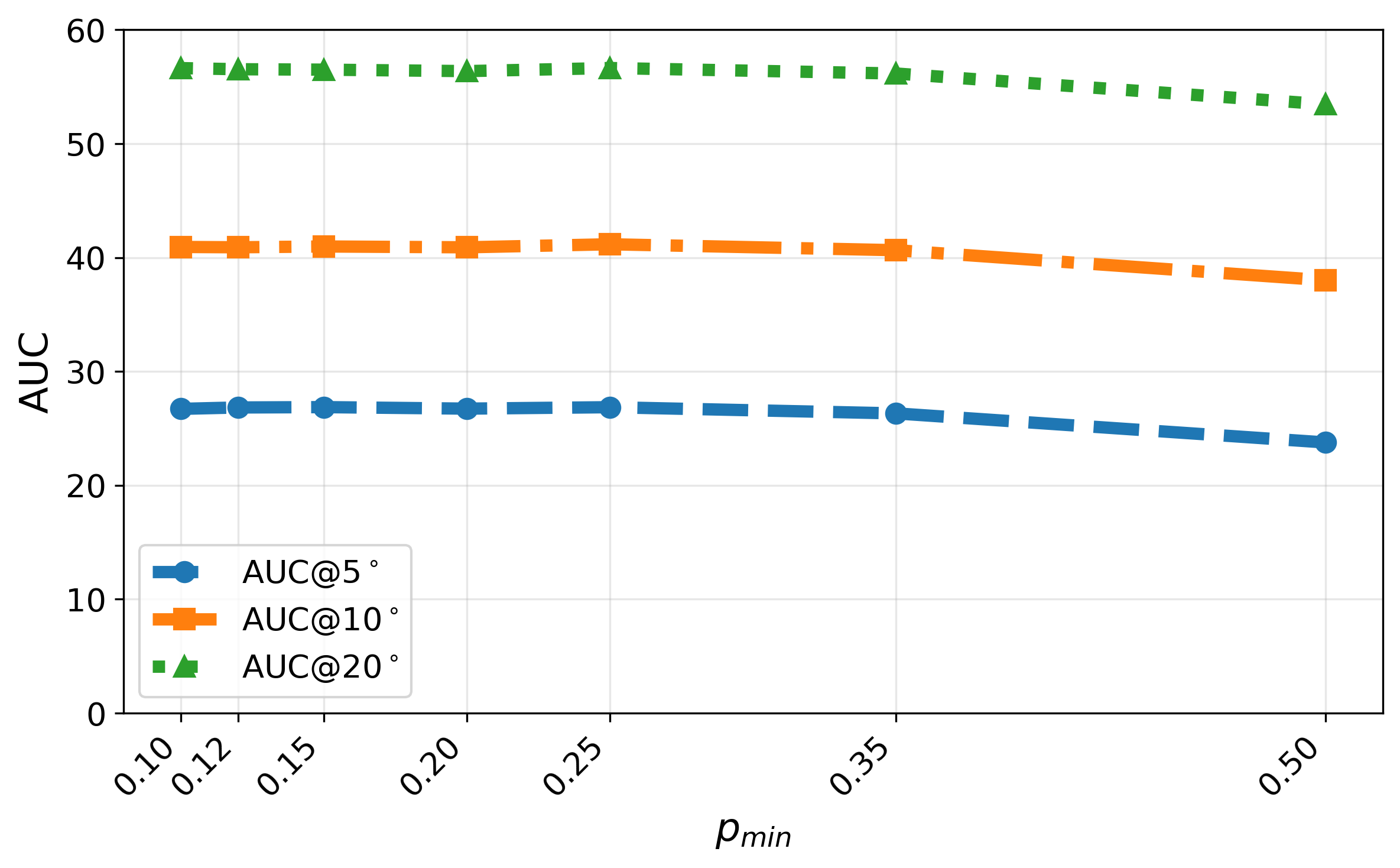}
        \caption{HM: AUC vs.\ minimum partition size \\ ($p_{\min}$)}
        \label{fig:fixed_pss_HM}
    \end{subfigure}
    \hfill
    \begin{subfigure}{0.48\linewidth}
        \centering
        \includegraphics[width=\linewidth]{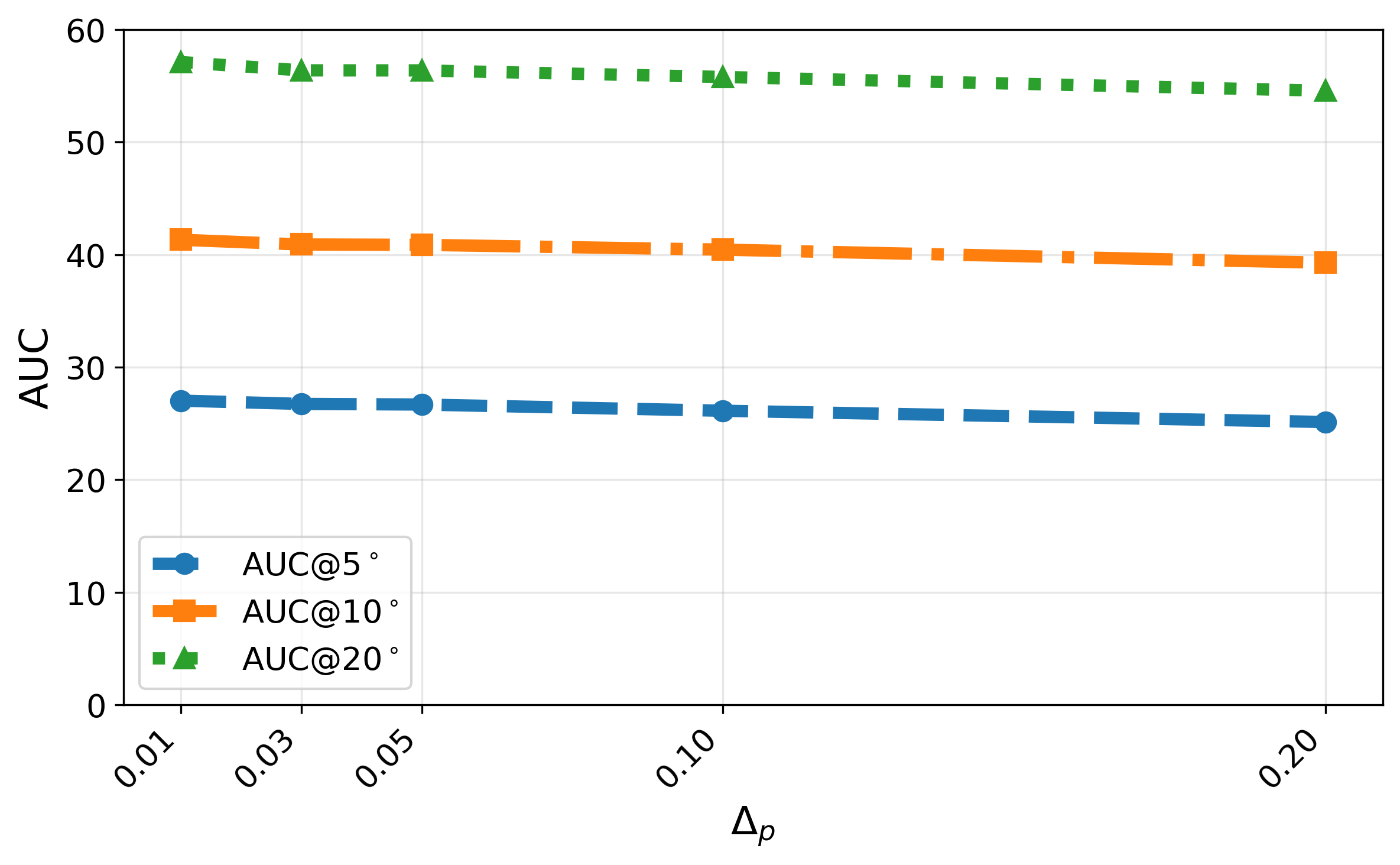}
        \caption{HM: AUC vs.\ percentile step size \\ ($\Delta p$)}
        \label{fig:fixed_mpf_HM}
    \end{subfigure}

    \vspace{0.35cm}

    \begin{subfigure}{0.48\linewidth}
        \centering
        \includegraphics[width=\linewidth]{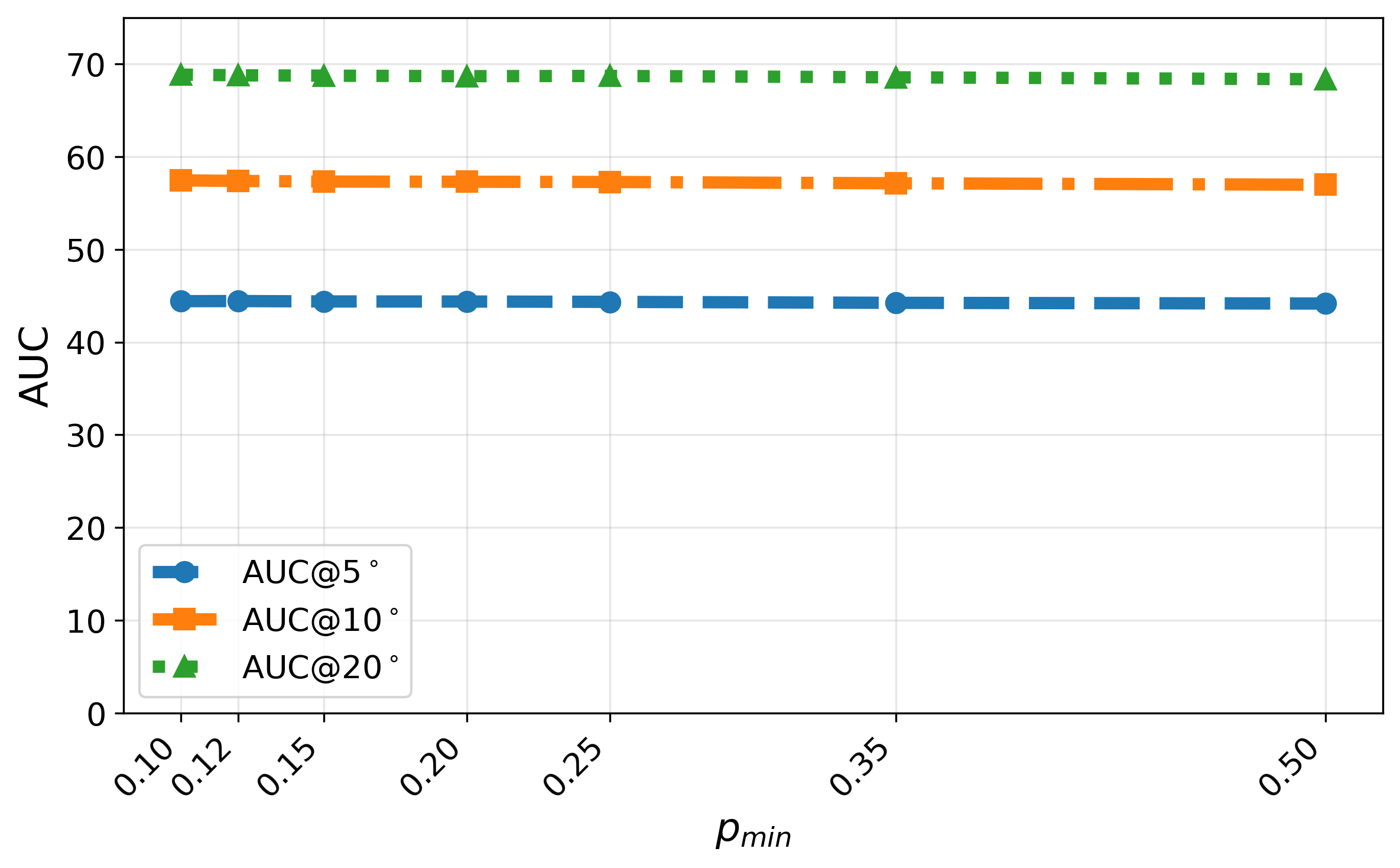}
        \caption{FM: AUC vs.\ minimum partition size \\ ($p_{\min}$)}
        \label{fig:fixed_pss_FM}
    \end{subfigure}
    \hfill
    \begin{subfigure}{0.48\linewidth}
        \centering
        \includegraphics[width=\linewidth]{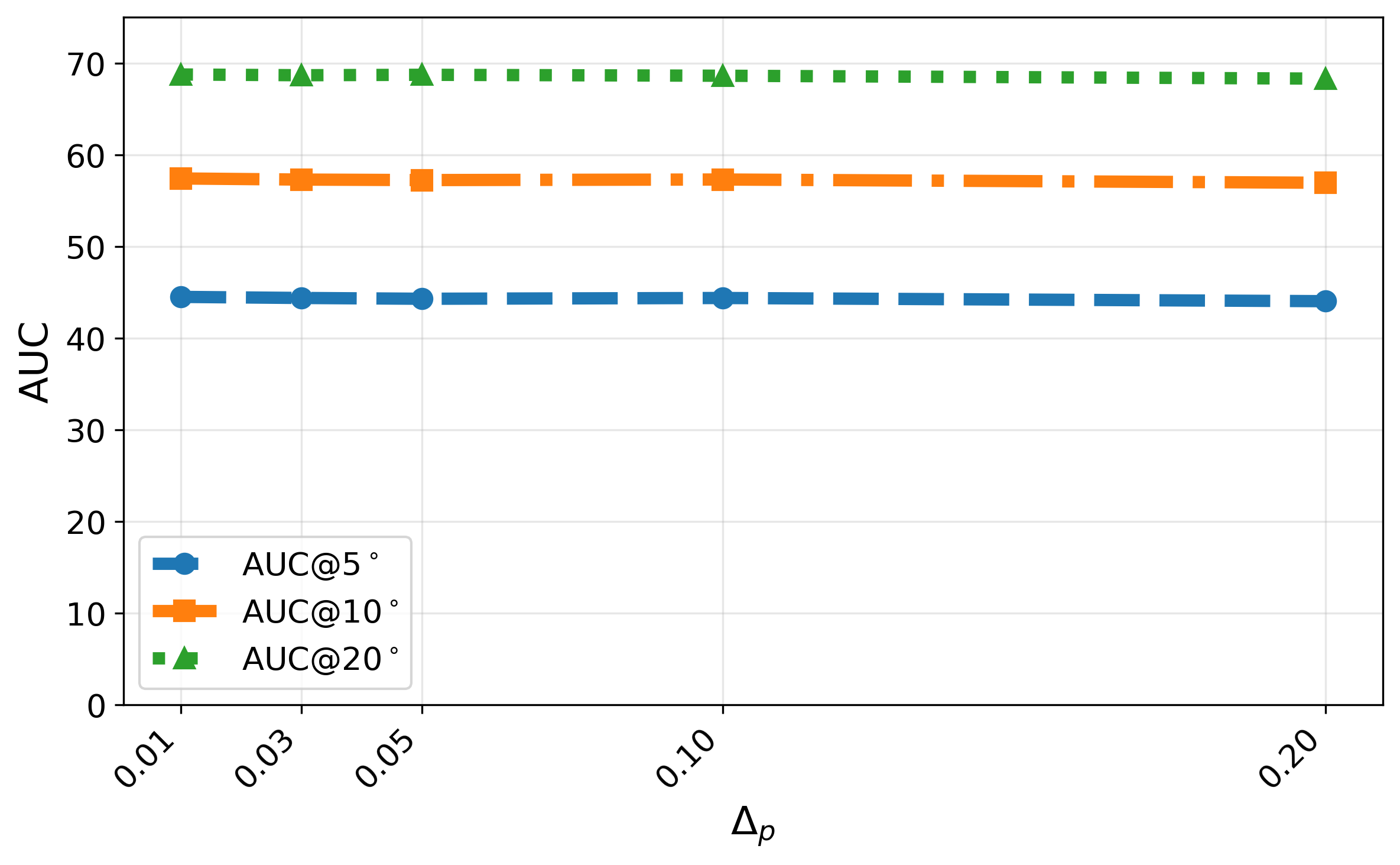}
        \caption{FM: AUC vs.\ percentile step size \\ ($\Delta p$)}
        \label{fig:fixed_mpf_FM}
    \end{subfigure}

    \vspace{0.35cm}

    \begin{subfigure}{0.48\linewidth}
        \centering
        \includegraphics[width=\linewidth]{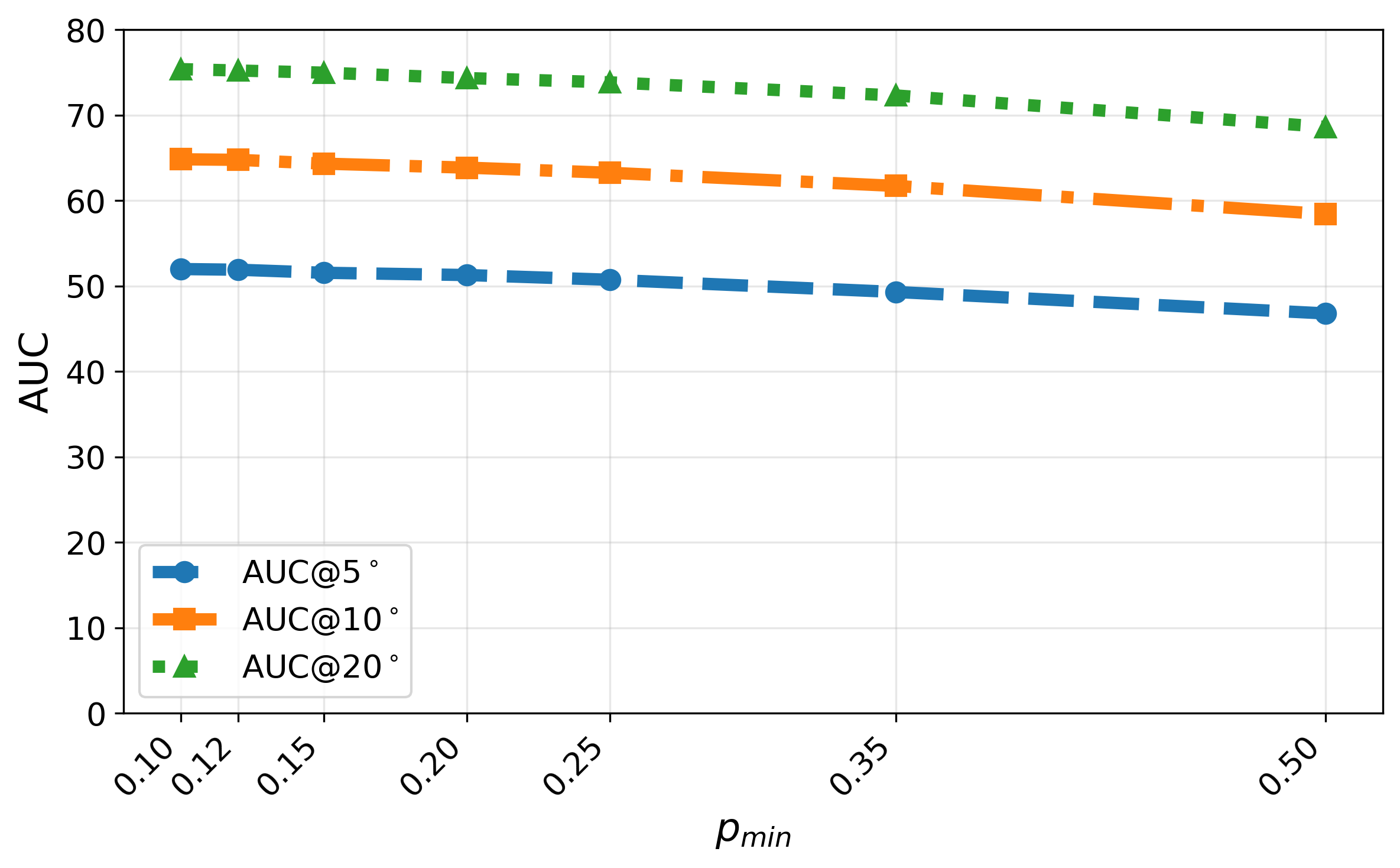}
        \caption{EM: AUC vs.\ minimum partition size \\ ($p_{\min}$)}
        \label{fig:fixed_pss_EM}
    \end{subfigure}
    \hfill
    \begin{subfigure}{0.48\linewidth}
        \centering
        \includegraphics[width=\linewidth]{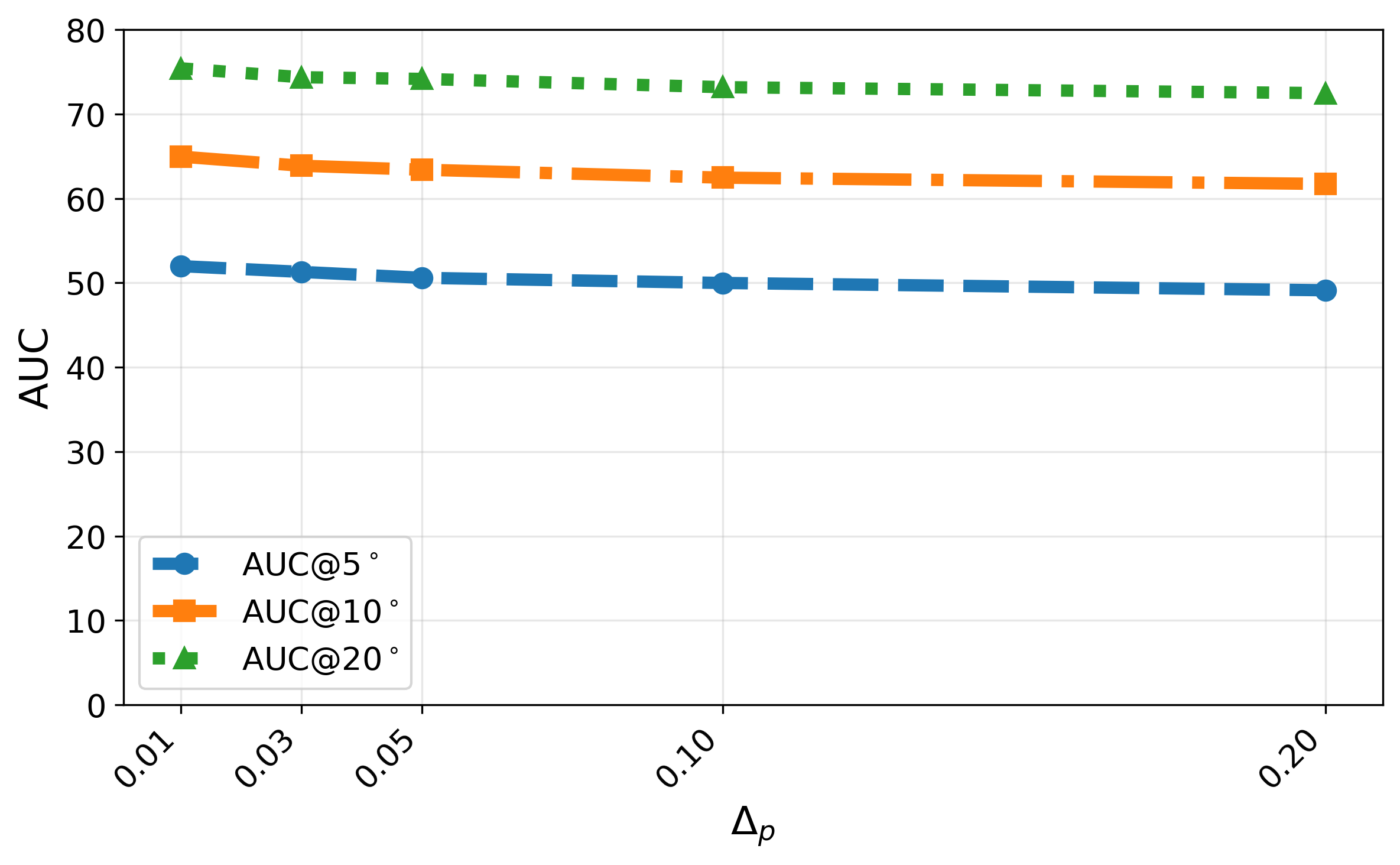}
        \caption{EM: AUC vs.\ percentile step size \\ ($\Delta p$)}
        \label{fig:fixed_mpf_EM}
    \end{subfigure}

    \caption{Sensitivity analysis of DS-SAC under different parameter settings. The first column shows the effect of varying the minimum partition size while keeping the percentile step size fixed, and the second column shows the effect of varying the percentile step size while keeping the minimum partition size fixed. Results are shown for homography (HM), fundamental matrix (FM), and essential matrix (EM) estimation.}
    \label{fig:sensitivity_analysis}
\end{figure}

Overall, DS-SAC consistently achieves the highest AUC scores across fundamental matrix, essential matrix, and homography estimation. The method also provides competitive or best median pose errors while maintaining the lowest runtime in all three experiments. These results demonstrate that DS-SAC is an accurate and efficient robust estimator for large-scale geometric vision tasks.

\subsection{Sensitivity Analysis}

From Fig.~\ref{fig:sensitivity_analysis}, we observe that DS-SAC is relatively stable under different choices of percentile step size and minimum partition size. When analyzing the effect of the minimum partition size, the percentile step size is fixed to $\Delta p=0.03$. Similarly, when analyzing the effect of the percentile step size, the minimum partition size is fixed to $p_{\min}=0.2$. In general, smaller values of these parameters allow the algorithm to search the residual space more finely, which can lead to slightly higher AUC scores. However, smaller values also increase the number of iterations and computational cost. Therefore, we select the final parameters by considering the trade-off among AUC, median error, and runtime. Based on this analysis, we set the minimum partition size to $p_{\min}=0.2$ and the percentile step size to $\Delta p=0.03$ in our experiments.

\begin{figure}[htbp]
    \centering

    \begin{subfigure}{0.48\linewidth}
        \centering
        \includegraphics[width=\linewidth]{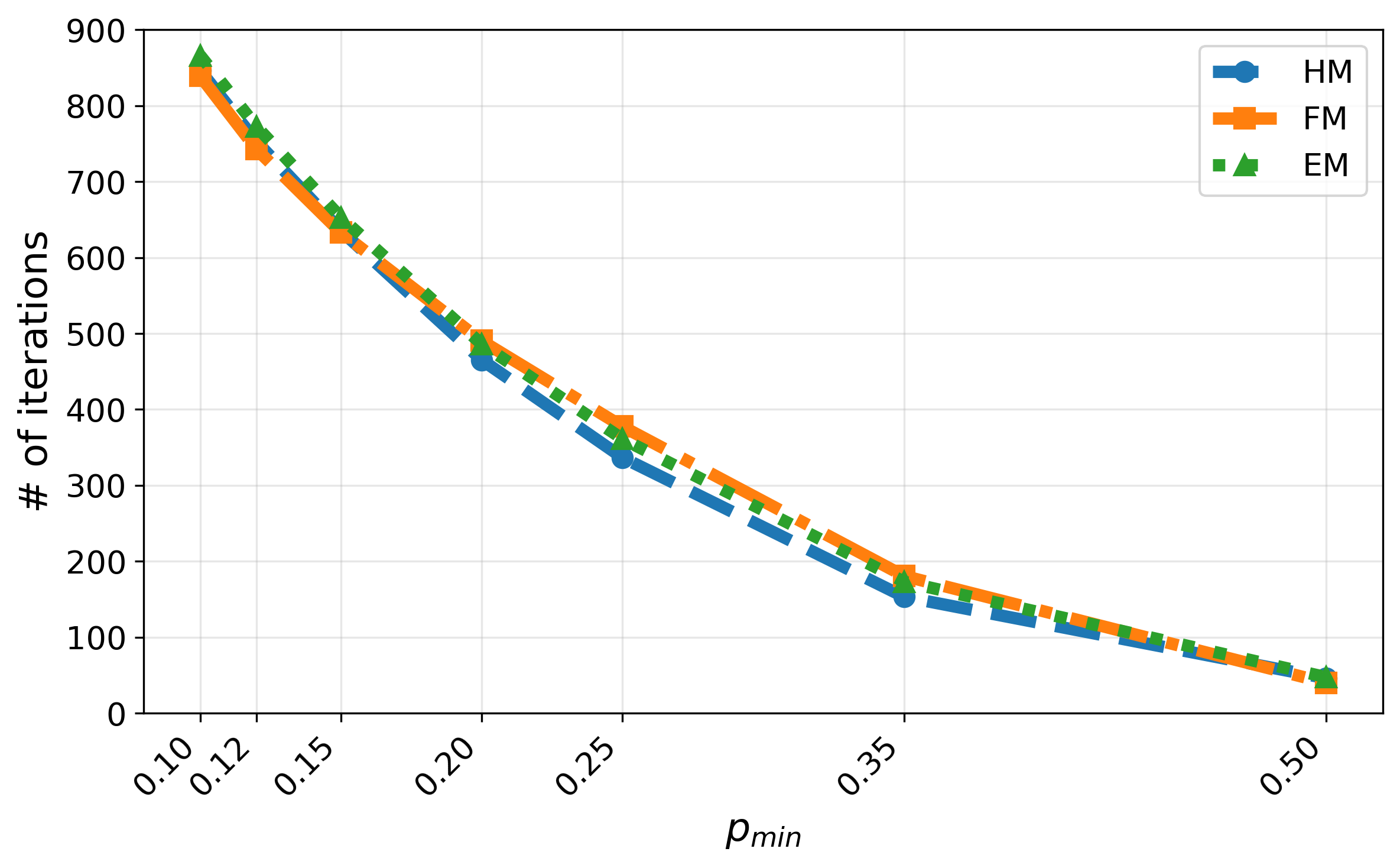}
        \caption{Number of iterations vs. minimum partition size ($p_{min}$)}
        \label{fig:iters_fixed_pss}
    \end{subfigure}
    \hfill
    \begin{subfigure}{0.48\linewidth}
        \centering
        \includegraphics[width=\linewidth]{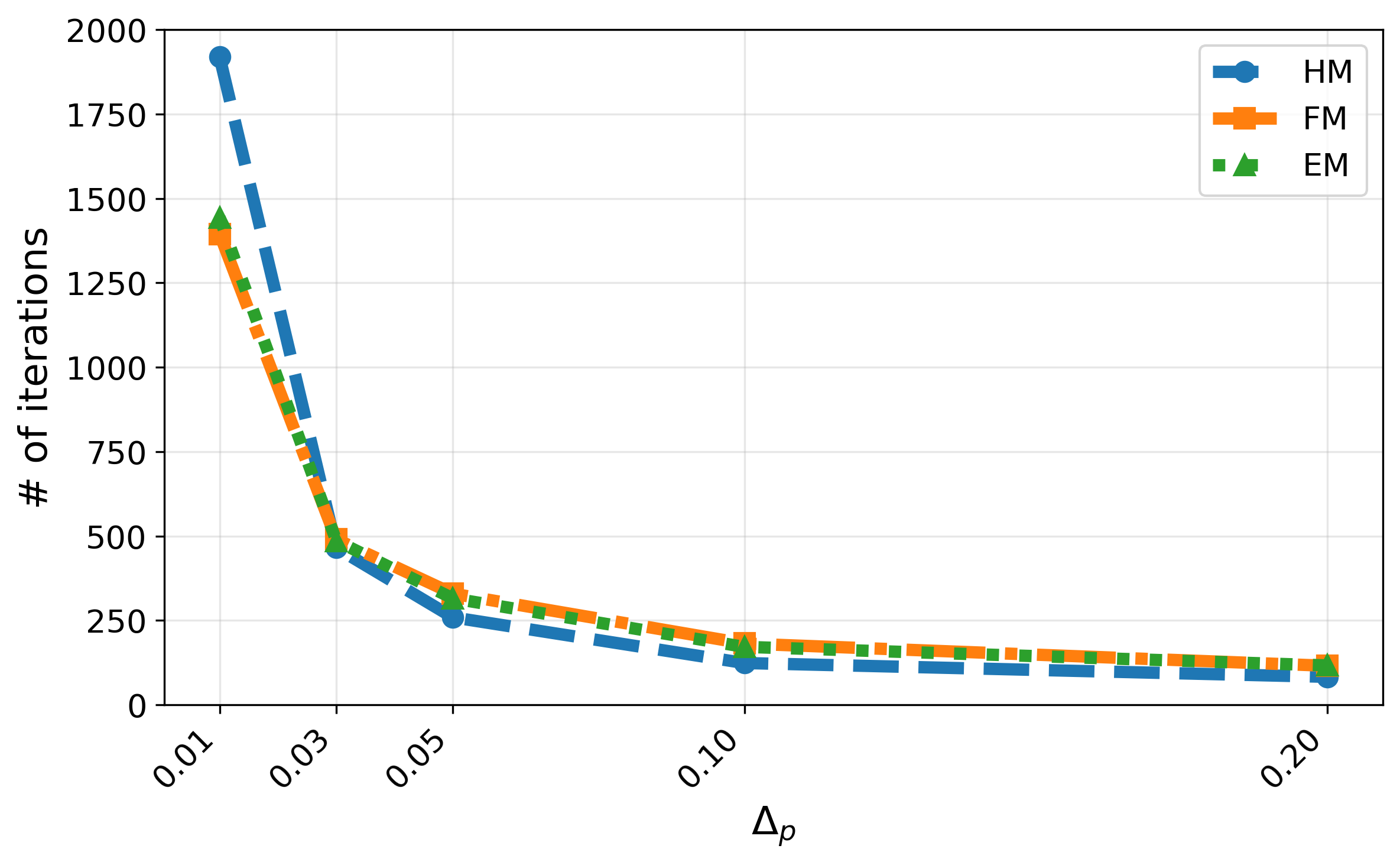}
        \caption{Number of iterations vs. percentile step size ($\Delta p$)}
        \label{fig:iters_fixed_mpf}
    \end{subfigure}

    \caption{Sensitivity analysis of DS-SAC with respect to the number of iterations. The left plot shows the effect of varying the minimum partition size, while the right plot shows the effect of varying the percentile step size.}
    
    \label{fig:iters_sensitivity_analysis}
\end{figure}
The sensitivity analysis shows that DS-SAC maintains good performance even when relatively large partition sizes are used. For example, the method remains stable for small percentile step sizes such as $\Delta p=0.01$ and for large minimum partition sizes such as $p_{\min}=0.5$. As shown in the Fig. \ref{fig:iters_sensitivity_analysis}, the number of iterations decreases almost linearly as the minimum partition size increases. Similarly, the number of iterations is inversely related to the percentile step size: larger values of $\Delta p$ lead to fewer iterations. Here, a minimum partition size of $p_{\min}=0.1$ means that each valid partition must contain at least $10\%$ of the total input points.

\subsection{Ablation Study}


Table~\ref{tab:comparison_table} analyzes the contribution of backward search and post-tuning in DS-SAC by disabling each component separately while keeping all other parameters unchanged. When post-tuning is disabled, the performance remains almost unchanged for homography estimation, but it decreases for fundamental and essential matrix estimation. In particular, the standard DS-SAC achieves higher AUC scores and lower median errors for fundamental and essential matrix estimation, with only a marginal increase in runtime. This indicates that post-tuning is more beneficial for epipolar geometry estimation than for homography estimation.

Disabling backward search reduces the computational cost across all three tasks, but it also leads to a drop in accuracy, especially for essential matrix estimation. For homography and essential matrix estimation, the standard DS-SAC achieves better AUC scores and lower median errors, showing the importance of backward search for improving model refinement. However, for fundamental matrix estimation, the performance without backward search remains very close to the standard DS-SAC while reducing the runtime from $0.008$s to $0.005$s. Therefore, using only forward search can be an efficient alternative for fundamental matrix estimation when runtime is more important than a small accuracy gain.

\begin{table}[htbp]
  \centering
  \begin{tabular}{>{\centering\arraybackslash}p{1.5cm}ccccccc}
    \toprule
    Setting & Task & AUC@$5^\circ$ & AUC@$10^\circ$ & AUC@$20^\circ$ & Med. Error & \# Inliers & Time (s) \\
    \midrule
    \multirow{3}{1.5cm}{\centering Standard DS-SAC}
    & HM & \textbf{26.74} & \textbf{40.90} & 56.38 & \textbf{3.87} & \textbf{55.68} & 0.007 \\
    & FM & \textbf{44.38} & \textbf{57.29} & \textbf{68.68} & \textbf{2.06} & \textbf{283.29} & 0.008 \\
    & EM & \textbf{51.29} & \textbf{63.87} & \textbf{74.35} & \textbf{1.72}  & \textbf{275.46} & 0.015 \\
    \midrule
    \multirow{3}{1.5cm}{\centering Without backward search}
    & HM & 26.29 & 40.44 & 56.00 & 4.05 & 54.67 & \textbf{0.004} \\
    & FM & 44.27 & 57.17 & 68.57 & 2.06 & 283.17 & \textbf{0.005} \\
    & EM & 50.16 & 62.76 & 73.41 & 1.86 & 273.93 & \textbf{0.007} \\
    \midrule
    \multirow{3}{1.5cm}{\centering Without post-tuning}
    & HM & \textbf{26.74} & \textbf{40.90} & \textbf{56.43} & \textbf{3.87} & 55.42 & 0.007 \\
    & FM & 43.60 & 56.79 & 68.36 & 2.11 & 281.04 & 0.008 \\
    & EM & 49.76 & 62.86 & 73.77 & 1.79 & 270.65 & 0.014 \\
    \bottomrule
  \end{tabular}
  \caption{Ablation study of backward search and post-tuning in DS-SAC. The standard DS-SAC configuration is compared with variants in which backward search and post-tuning are disabled separately.}
  \label{tab:comparison_table}
\end{table}

\section{Conclusion}\label{sec5}

In this work, we introduced DS-SAC, a deterministic framework for robust geometric model estimation. Unlike RANSAC-based methods that rely on stochastic minimal sampling, DS-SAC uses all available points to initialize the model and then systematically refines it by searching dense regions. The proposed forward and backward search strategy, combined with recursive residual-based partitioning, enables DS-SAC to explore promising model regions while reducing unnecessary random hypothesis generation. Through extensive experiments on large-scale real-world datasets for homography, fundamental matrix, and essential matrix estimation, DS-SAC consistently achieved higher AUC scores, competitive or lower median pose errors, and faster runtime compared with widely used robust estimators. These results demonstrate that DS-SAC provides an effective and efficient alternative to stochastic consensus-based methods for geometric vision tasks.

\section*{Data Availability Declaration}
The datasets used in this work are publicly available. The PhotoTourism dataset is available from the CVPR tutorial ``RANSAC in 2020'' at \url{http://cmp.felk.cvut.cz/~mishkdmy/CVPR-RANSAC-Tutorial-2020/RANSAC-Tutorial-Data-EF.tar}. The ScanNet test set for relative pose estimation used in SuperGlue is available at \url{https://www.polybox.ethz.ch/index.php/s/lAZyxm62WUh27Zl/download}. The 7Scenes RGB-D dataset is available at \url{https://www.microsoft.com/en-us/research/project/rgb-d-dataset-7-scenes/}. The ETH3D high-resolution multi-view dataset is available at \url{https://www.eth3d.net/datasets}. The LaMAR CAB scene is available at \url{https://cvg-data.inf.ethz.ch/lamar/CAB.zip}. The KITTI Visual Odometry dataset is available at \url{https://www.cvlibs.net/datasets/kitti/eval_odometry.php}.

\bibliography{sn-bibliography}
\end{document}